\documentclass[10pt,twocolumn,letterpaper]{article}

\usepackage{cvpr}              % To produce the CAMERA-READY version
\usepackage{minted}
\usepackage[breakable, skins]{tcolorbox}
\definecolor{codeBackgroundColor}{RGB}{248, 248, 248}
\tcbset{
    colback=codeBackgroundColor, frame empty, opacityframe=0,
    boxsep=2mm, left=0mm, right=0mm, top=0mm, bottom=0mm,
    rounded corners=all, arc=5pt, breakable
}
\newenvironment{code*}[1]{
    \VerbatimEnvironment \begin{tcolorbox} \begin{minted}[
        style=default, fontsize=\small, linenos,
        baselinestretch=1.3, breaklines]{#1}}
{\end{minted} \end{tcolorbox} }
\newenvironment{codenn*}[1]{
    \VerbatimEnvironment \begin{tcolorbox} \begin{minted}[
        style=default, fontsize=\small,
        baselinestretch=1.3, breaklines]{#1}}
{\end{minted} \end{tcolorbox} }
\definecolor{cvprblue}{rgb}{0.21,0.49,0.74}
\usepackage[pagebackref,breaklinks,colorlinks,allcolors=cvprblue]{hyperref}

\def\paperID{} % *** Enter the Paper ID here
\def\confName{CVPR}
\def\confYear{2026}
\def\modelname{PROMO\xspace}
\title{PROMO: Promptable Outfitting for Efficient High-Fidelity Virtual Try-On}

\author{
Haohua Chen\textsuperscript{1,2,*} \quad
Tianze Zhou\textsuperscript{1,3,*} \quad
Wei Zhu\textsuperscript{1} \quad
Runqi Wang\textsuperscript{1} \quad
Yandong Guan\textsuperscript{1,2} \\
Dejia Song\textsuperscript{1} \quad
Yibo Chen\textsuperscript{1} \quad
Xu Tang\textsuperscript{1} \quad
Yao Hu\textsuperscript{1} \quad
Lu Sheng\textsuperscript{2,$\dagger$} \quad
Zhiyong Wu\textsuperscript{3,$\ddagger$} \\
\textsuperscript{1}Xiaohongshu Inc. \qquad 
\textsuperscript{2}Beihang University \qquad 
\textsuperscript{3}Tsinghua University \\
}
\usepackage{graphicx}
\usepackage{multirow}
\begin{document}
\twocolumn[{%
\renewcommand\twocolumn[1][]{#1}%
\maketitle
\begin{center}
    \centering
    \captionsetup{type=figure}
    \vspace{-3mm}
\includegraphics[width=0.85\textwidth]{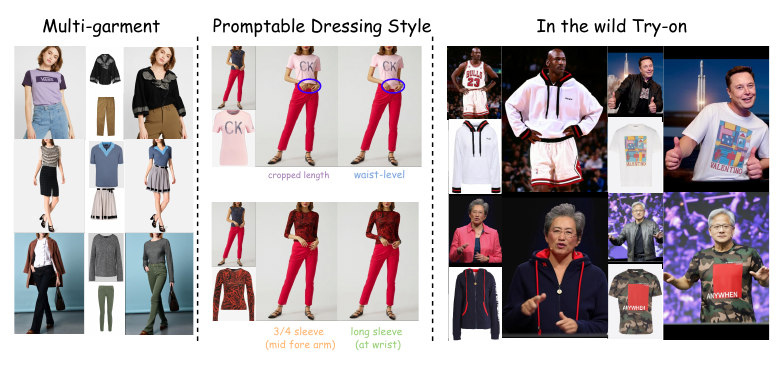}
    \vspace{-5mm}
    \captionof{figure}{\modelname enables multi-garment try-on, prompt-based control over dressing styles and demonstrates robust performance in challenging real-world scenarios.
    }
    \label{fig:abs}
\end{center}%
}]

\maketitle
\let\thefootnote\relax\footnotetext{\noindent * Equal contribution.\\
$^{\dagger}$ Corresponding author: lsheng@buaa.edu.cn.\\
$^{\ddagger}$ Corresponding author: zywu@sz.tsinghua.edu.cn}
\begin{abstract}
Virtual Try-on (VTON) has become a core capability for online retail, where realistic try-on results provide reliable fit guidance, reduce returns, and benefit both consumers and merchants. Diffusion-based VTON methods achieve photorealistic synthesis, yet often rely on intricate architectures such as auxiliary reference networks and suffer from slow sampling, making the trade-off between fidelity and efficiency a persistent challenge.
We approach VTON as a structured image editing problem that demands strong conditional generation under three key requirements: subject preservation, faithful texture transfer, and seamless harmonization. Under this perspective, our training framework is generic and transfers to broader image editing tasks. Moreover, the paired data produced by VTON constitutes a rich supervisory resource for training general-purpose editors.
We present \modelname, a promptable virtual try-on framework built upon a Flow Matching DiT backbone with latent multi-modal conditional concatenation. By leveraging conditioning efficiency and self-reference mechanisms, our approach substantially reduces inference overhead. On standard benchmarks, \modelname surpasses both prior VTON methods and general image editing models in visual fidelity while delivering a competitive balance between quality and speed. These results demonstrate that flow-matching transformers, coupled with latent multi-modal conditioning and self-reference acceleration, offer an effective and training-efficient solution for high-quality virtual try-on.
\end{abstract}    
\section{Introduction}
\label{sec:intro}

Virtual Try-on (VTON) aims to render a target garment onto a given person image with high fidelity and realism. It is of practical importance in e-commerce, where it enables shoppers to obtain reliable fit and appearance references without in-store trials, provides interactive experiences, and reduces return rates, benefiting both consumers and merchants. 

Early VTON systems predominantly relied on warping-based pipelines~\cite{he2022style,zhao2022thin},. These approaches often yield unnatural appearances and degrade under challenging poses, occlusions, and large non-rigid deformations.

GAN methods~\cite{men2020controllableADGAN,xie2022pastaganversatileframeworkhighresolution,lewis2021tryonganbodyawaretryonlayered,yang2020towards,VITON-gan}, improve realism to some extent, yet they commonly struggle to preserve fine garment details and textures, produce coherent shading and illumination, and maintain natural human geometry and articulation. 

Spurred by the superior image synthesis performance of diffusion models~\cite{ho2020denoisingdiffusionprobabilisticmodels,rombach2021highresolution,peebles2023scalable}, a growing body of VTON research leverages diffusion-based generative modeling~\cite{Parts2Whole,chong2024catvtonconcatenationneedvirtual,xu2024ootdiffusion,guo2025any2anytryonleveragingadaptiveposition,morelli2023ladi,jiang2024fitditadvancingauthenticgarment,chong2025fastfitacceleratingmultireferencevirtual,kim2024promptdresserimprovingqualitycontrollability,choi2024improvingdiffusionmodelsauthenticidmvton,shen2024imagdressingv1customizablevirtualdressing}, demonstrating clear gains in visual fidelity and controllability. Many approaches use a diffusion UNet~\cite{ronneberger2015unetconvolutionalnetworksbiomedical} or Transformer model~\cite{vaswani2023attentionneed}, adding conditions by using a reference net to create condition tokens, which are then appended to the main denoising net’s key/query sequences. These methods~\cite{jiang2024fitditadvancingauthenticgarment,choi2024improvingdiffusionmodelsauthenticidmvton,xu2024ootdiffusion,shen2024imagdressingv1customizablevirtualdressing}, require another whole net, resulting in complex initialization and interaction logic between the two nets. Other approaches~\cite{chong2024catvtonconcatenationneedvirtual,ni2025itvtonvirtualtryondiffusion} use image-level concatenation (e.g., concatenating condition images as a larger image) to inject conditions. However, considering information density, parts of the conditions do not need to be at the same resolution as the output images.

Moreover, the dressing style of the garment varies. Some works overlook this, replacing the text encoder with an image encoder. PromptDresser~\cite{kim2024promptdresserimprovingqualitycontrollability} uses prompts to control dressing style, but it requires a closed-source VLM to caption the dressing style.

In summary, the contributions of this work include:
\begin{itemize}[topsep=3pt, partopsep=3pt,leftmargin=5pt, itemsep=3pt]
\item We propose \modelname, a multi-modal, prompt-controllable (promptable) virtual try-on framework. Leveraging a temporal self-reference mechanism across denoise timesteps, group-wise attention mask modulation, and efficient token merging, \modelname attains fast yet high-quality generation with a large-parameter Diffusion Transformer (DiT) backbone.
\item Motivated by real-world virtual try-on applications, we propose a Style Prompt System by tuning Qwen2.5-VL-7B, which provides reliable style description to support promptable control and downstream evaluation.
\item We conduct extensive experiments on the public VITON-HD and DressCode datasets and demonstrate consistent gains over state-of-the-art baselines VTON models and image-editing model in benchmark, achieving superior visual quality and realism.
\end{itemize}

\begin{figure*}[h!]
    \centering
    \includegraphics[width=0.90\textwidth]{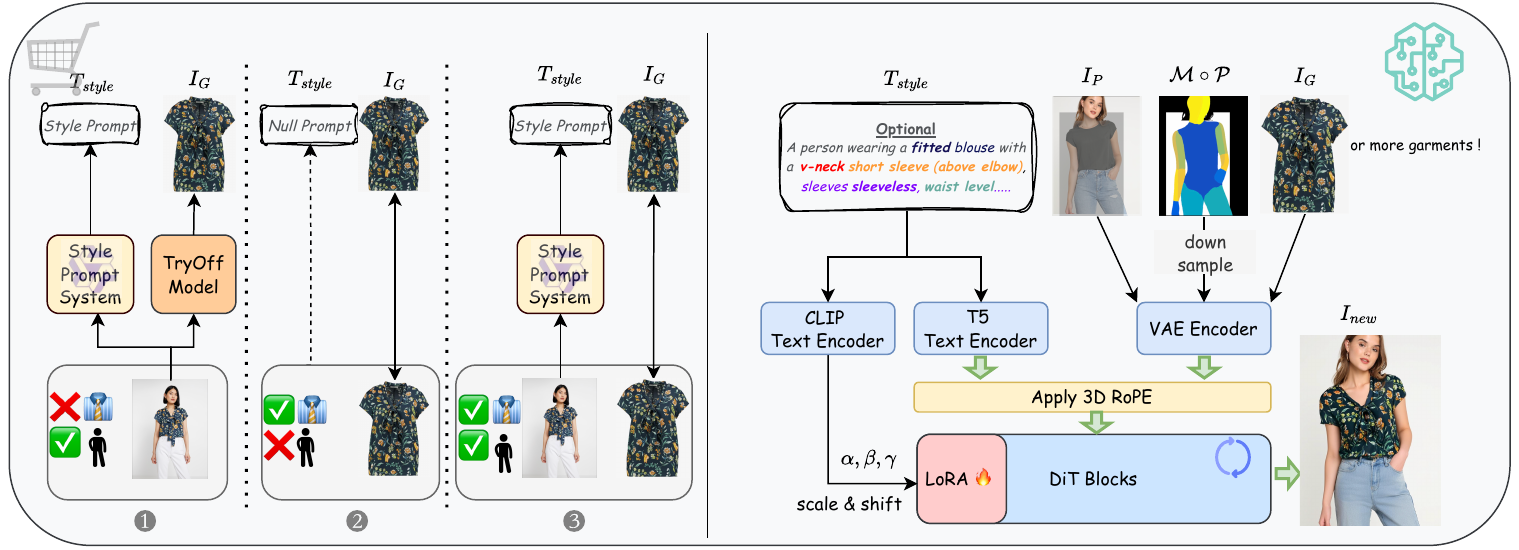} 
    \vspace{-2mm}
    \caption{Overview of scenarios handled by our \modelname framework. The left panel illustrates three common scenarios that customers encounter during online shopping: 1) model image available without a specific garment image, 2) garment image available without a model, and  3) both model and garment images available. Our system addresses the missing information in each scenario by generating the necessary conditional inputs required by our model. Notably, for scenarios without model images, our system features pure image-reference capability, allowing text prompts to be omitted while maintaining a robust baseline performance.}
    \label{fig:model}
    \vspace{-5mm}
\end{figure*}

\section{Related Work}
\label{sec:related_work}

%-------------------------------------------------------------------------
\subsection{Image-based Virtual Try-on}

Image-based virtual try-on synthesizes realistic images of a person wearing a target garment. Early warping-based methods used a two-stage pipeline (warping and fusion) with techniques like Thin Plate Spline (TPS)~\cite{zheng2019virtually,yang2020towards} or appearance flow~\cite{he2022style}, but suffered from alignment artifacts.

Diffusion-based approaches now dominate, avoiding explicit warping. LaDI-VTON~\cite{morelli2023ladi} and StableVITON~\cite{kim2024stableviton} use ControlNet~\cite{zhang2023adding}-inspired conditioning. Others, like OOTDiffusion~\cite{xu2024ootdiffusion}, IMAGDressing-v1~\cite{shen2025imagdressing} and IDM-VTON~\cite{choi2024improvingdiffusionmodelsauthenticidmvton}, employ parallel ``reference nets" to encode garment features.

Recent works like PromptDresser~\citep{kim2024promptdresserimprovingqualitycontrollability} leverages LMMs for text-editable style control (\eg``tuck", ``fit") via prompt-aware masks to enhance controllability. Any2AnyTryon~\citep{guo2025any2anytryonleveragingadaptiveposition} uses a unified, mask-free DiT framework for multiple tasks, like layered try-on and garment reconstruction , using adaptive position embeddings to handle variable inputs.

Existing diffusion methods often require complex architectures, motivating our pursuit of a more efficient framework.

\subsection{Diffusion Models}

% U-Net\citep{ronneberger2015unetconvolutionalnetworksbiomedical} has long served as the predominant backbone architecture for diffusion-based image synthesis, widely adopted in text-to-image generation models including LDM~\citep{rombach2021highresolution}, SDXL~\citep{podell2024sdxl}, DALL-E~\citep{ramesh2021zero}, and Imagen~\citep{saharia2022photorealistic}. DiT~\citep{peebles2023scalable} pioneered the replacement of U-Net with Transformer architectures for class-conditional image generation, demonstrating a strong correlation between network capacity and generation quality.

% Building upon this foundation, Stable Diffusion 3~\citep{esser2024scalingrectifiedflowtransformerssd3} and Flux introduce transformer-based frameworks employing MM-DiT blocks, which utilize separate weight parameters for text and image modalities to achieve remarkable text-to-image synthesis performance. These models adopt flow-matching formulations to predict the velocity field from noise to clean images.

% In this work, we leverage the Diffusion Transformer architecture due to its highly scalable nature, making it particularly well-suited as the foundational backbone for high-quality virtual try-on image generation. The inherent scalability of transformers enables our model to effectively capture fine-grained garment details and style control.

U-Net~\citep{ronneberger2015unetconvolutionalnetworksbiomedical} was the predominant backbone for diffusion models like LDM~\citep{rombach2021highresolution} and SDXL~\citep{podell2024sdxl}. DiT~\citep{peebles2023scalable} successfully replaced the U-Net with a Transformer, demonstrating strong scalability. Recent models like Stable Diffusion 3~\citep{esser2024scalingrectifiedflowtransformerssd3} and FLUX~\cite{flux2024,labs2025flux1kontextflowmatching} build on this, using transformer blocks and flow-matching formulations to predict velocity fields. We leverage the Diffusion Transformer architecture for its scalability, making it an ideal backbone for high-quality virtual try-on and capturing fine-grained garment details.

\subsection{Multi-Condition Image Generation}
Virtual try-on is essentially a multi-conditional portrait generation task, closely related to multi-condition image generation models. FLUX-Kontext~\cite{labs2025flux1kontextflowmatching} extends FLUX~\cite{labs2025flux1kontextflowmatching} to support text-plus-single-image conditioning, but struggles when handling multiple reference images. DreamO~\cite{mou2025dreamo} injects conditions via learnable embeddings, which suffer from limited generalization. UNO~\cite{wu2025lessuno}proposes a model-data co-evolution paradigm that leverages Text-to-Image (T2I) models to generate high-quality single-subject data, thereby training stronger Subject-to-Image (S2I) models capable of producing diverse multi-subject outputs. OminiControl~\cite{tan2025ominicontrol} separates subject and spatial conditions by toggling 2D positional-encoding offsets, implicitly assuming one reference per condition type, an assumption that fails in try-on scenarios with multiple garments. IC-LoRA~\cite{huang2024incontextloradiffusiontransformers} concatenates images at pixel level, ignoring heterogeneous information density across conditions and either sacrificing resolution or increasing computational cost.

\section{Method}

\subsection{Overview}
\subsubsection{Modeling}
The overall framework of \modelname is shown in \cref{fig:model}. Given a person image $I_{P}$, garment images $I_{G_{i}}$ and optional dressing style description $T_{style}$ extracted by Style Prompt System from $I_{model}$. Our model aims to generate a image $I_{n ew}$, where the person wears the given garments in the specified dressing style. It can be formulated as:
\begin{equation}
  I_{new} = \mathcal{F}(I_p, \{I_{G_i}\}_{i=1}^{N}, T_{style}^*)
\end{equation}
\begin{equation}
  \mathcal{C} = \{\mathcal{E}(\tilde{I}_p), \{\mathcal{E}(I_{G_i})\}_{i=1}^{N}, \mathcal{E}(\mathcal{M} \circ \mathcal{P})\}
\end{equation}
\begin{equation}
  \mathbf{z}_0 = \text{FM}(\mathbf{z}_T, \mathcal{C}, T_{style}^*), \quad \mathbf{z}_T \sim \mathcal{N}(0, \mathbf{I})
\end{equation}
\begin{equation}
  I_{new} = \mathcal{D}(\mathbf{z}_0)
\end{equation}
where $\tilde{I}_P = I_P \odot (1-\mathcal{M})$ is the masked person image, $\mathcal{E}$ is the unified image encoder, $\mathcal{M}$ and $\mathcal{P}$ denote segmentation masks and pose features with $\circ$ representing the overlay merging operation, $\text{FM}(\cdot)$ represents the flow matching model, $\mathcal{D}$ is the VAE decoder, and $^*$ indicates optional input.

\subsubsection{System for Real-World Scenarios}

Real-world e-commerce scenarios often present incomplete inputs. We address this with two auxiliary modules:

\noindent\textbf{Style Prompt System:} Generates textual descriptions $T_{style}$ of dressing styles from garment catalog images using a distilled Qwen VL 7B model (\cref{sec:promptable}).

\noindent\textbf{TryOff Model:} When an isolated garment image $I_{G_i}$ is unavailable,
we instead use the corresponding human model image $I_{\text{model}}$.
We extract the garment via our $\text{TryOff}$ module:
$$I_{G_i} = \text{TryOff}(I_{\text{model}}, T_{\text{garment}})$$
where $T_{\text{garment}}$ specifies the target garment region.
This enables the system to leverage un-paired data
during both training and inference, broadening its applicability.

\subsection{Precise and Efficient Spatial Conditioning}
\subsubsection{Agnostic Mask Generation}
Virtual try-on requires erasing clothing at corresponding body regions to prevent information leakage. Unlike traditional masks in datasets like VITON-HD~\cite{choi2021vitonhdhighresolutionvirtualtryon} that rely on body parsing with dilation or manual annotation, we adopt the approach from FitDiT~\cite{jiang2024fitditadvancingauthenticgarment}, combining human body parsing with DWPose~\cite{yang2023effectivedwpose} for automated mask generation. This method automatically generates upper-body, lower-body, or full-body masks based on the garment type, seamlessly integrating into our generation pipeline.

\subsubsection{Precise Body Shape Estimation}
% 在抹除原衣信息中作用中，mask具有必要性，然后会不可避免的擦除关于人体原本的体型和姿态，因此，提取人体的姿态和体型是实现高保真度虚拟试衣的关键一步。目前DensePose是一个人体部位和姿态和体型估计的模型，然而其在处理长裙、喇叭裤等宽松衣物等人像时，往往会导致其在肢体和躯干上的估计失真，在训练中会给模型带来信息泄漏，导致推理效果不佳。而DWPose是一个与体型无关，仅提取姿态的模型，然而，失去了体型信息，训练虚拟试衣模型在处理不同体型的人物时往往会无法实现准确的估计。
% 因此我们基于EOMT, 使用迭代式的图像生成，训练循环，生成了更加准确，无视覆盖衣物的姿态体型估计模型，将其整合进入我们的pipeline
Masks remove original clothing information but inevitably lose body shape and pose details. Therefore, extracting human pose and shape is critical for high-fidelity virtual try-on. DensePose~\cite{güler2018denseposedensehumanpose} estimates body parts, pose, and shape, but produces distorted results on loose-fitting garments like long skirts, causing information leakage and poor inference. DWPose~\cite{yang2023effectivedwpose} extracts only pose information without shape. However, lacking body shape data, pose-only models fail to handle different body types accurately.
\begin{figure}[tb]
    \centering
    \includegraphics[width=0.70\linewidth]{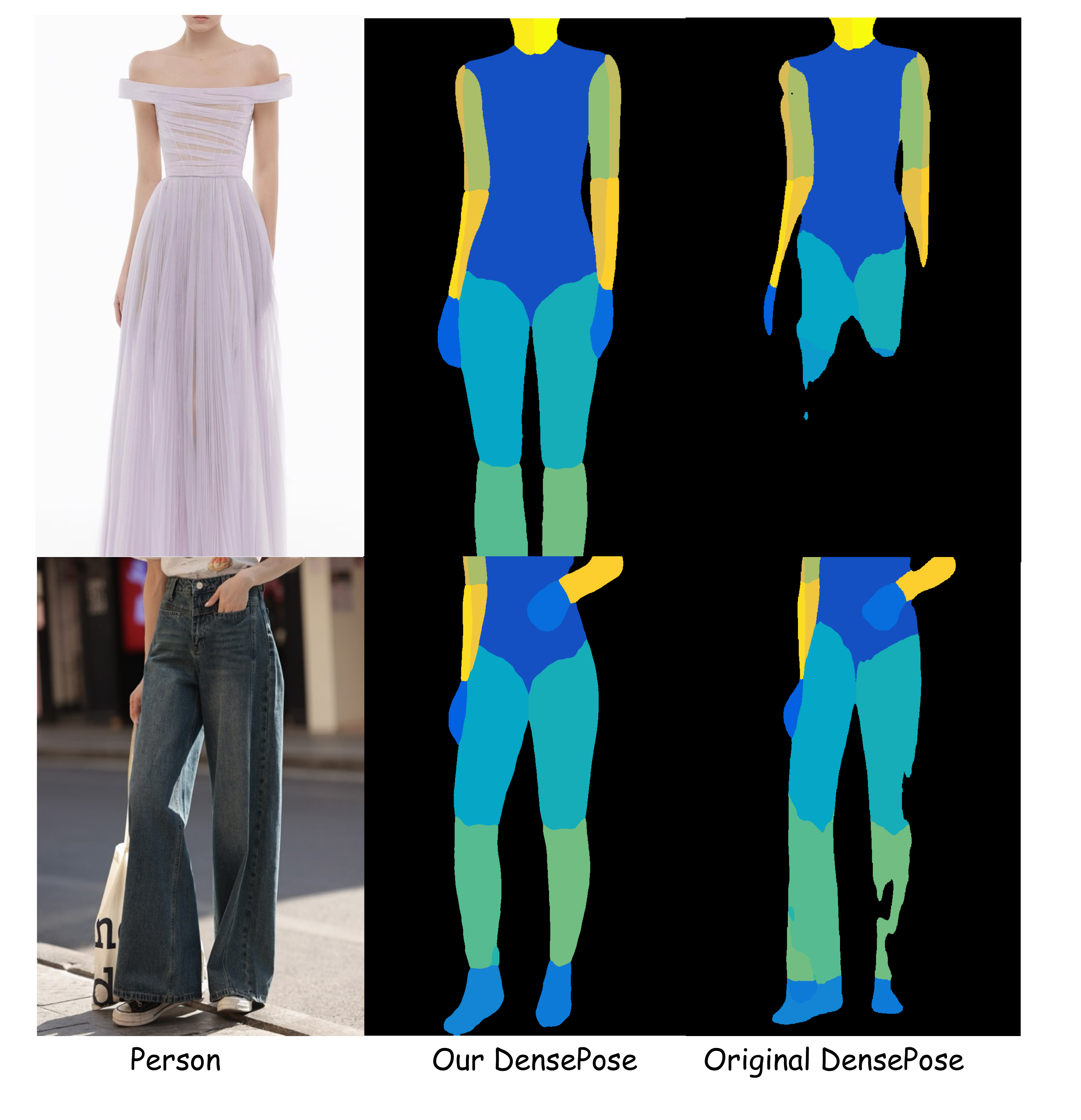}
    \vspace{-5mm}
    \caption{Compared to directly using the original DensePose~\cite{güler2018denseposedensehumanpose}, our method better estimates plausible body shapes under loose clothing, effectively preventing information leakage.}
    \vspace{-5mm}
    \label{fig:densepose}
\end{figure}
To address this, we leverage EOMT~\cite{kerssies2025eomt} and employ an iterative image generation training to develop a more accurate pose and shape estimation model that is robust to clothing occlusion. \cref{fig:densepose} shows the comparison of ours and original DensePose.
\begin{figure}[tb]
    \centering
    \includegraphics[width=1\linewidth]{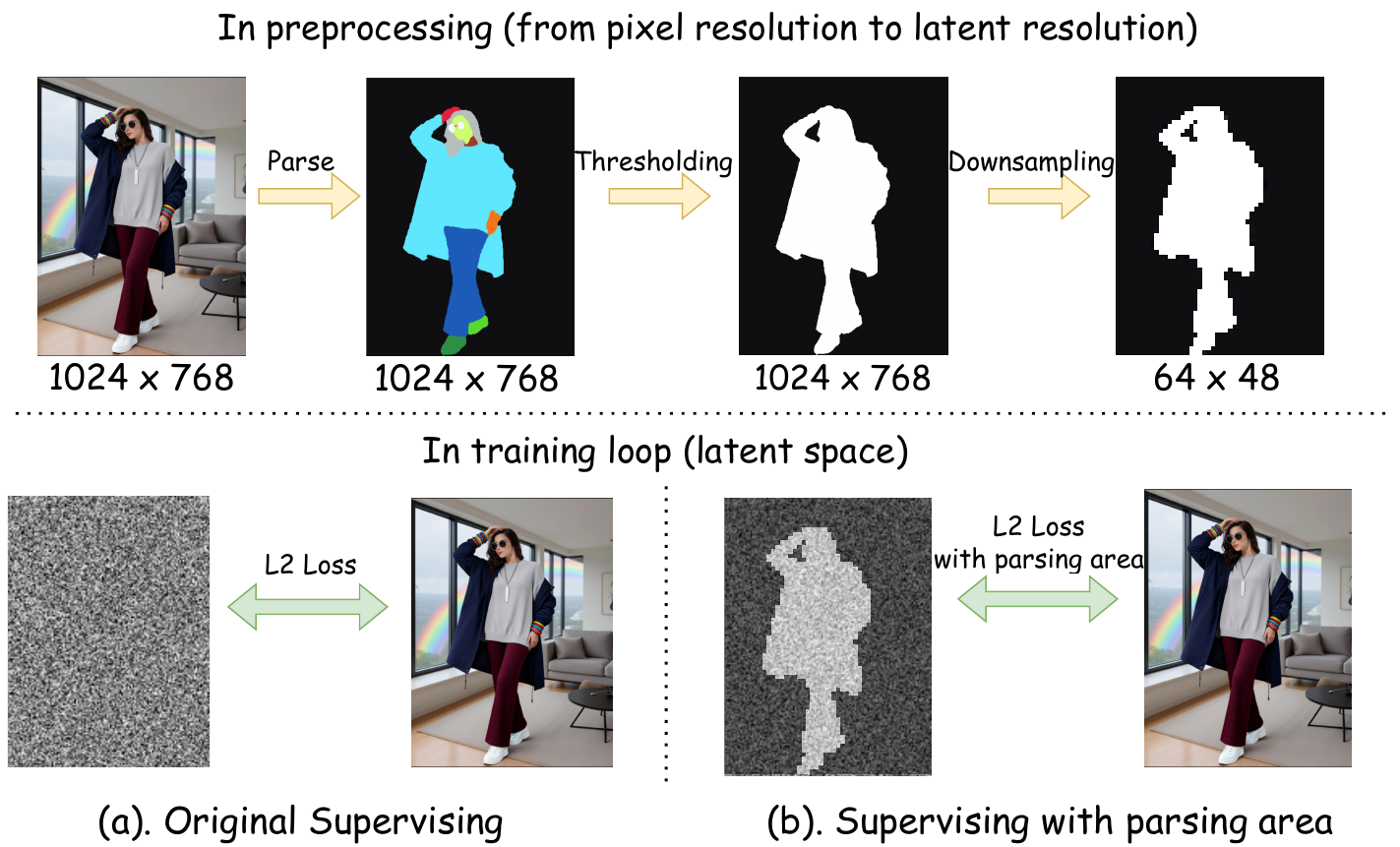}
    \vspace{-6mm}
    \caption{We downsample the extracted human parsing to match the resolution of the latent space. Compared to the standard supervision in (a), we adopt a weighted loss design in (b).}
    \label{fig:parsing_area}
    \vspace{-5mm}
\end{figure}
\subsubsection{Spatial Condition Merging}
\label{subsubsec:merge}
% IC-Lora, CatVton 对于图像条件注入的尝试，基于image级别的拼接，这需要拼接图片的分辨率保持一致，
% 在我们虚拟试衣的框架中，person的分辨率时1024x768，然而，mask和pose其实存在信息冗余，并不需要那么高的分辨率
% 收到ominicontrol2 的启发，我们在将mask以及pose条件送入VAE 进行encoder之前，对其长宽进行等比例的2倍的下采样，在latent空间里将t对应的oken数量变为原本的25%。基于对空间条件的信息冗余性分析，我们再提出，将pose条件粘贴在mask条件之上，至此，我们使得最后的merged条件，相较于最初的双条件，token数量压缩至原本的12.5%， 如图所示。这使得我们模型在推理和训练上都能更加高效
While IC-LoRA~\cite{huang2024incontextloradiffusiontransformers} and CatVTON~\cite{chong2024catvtonconcatenationneedvirtual} attempt to inject image conditions via image-level concatenation, this approach necessitates uniform resolution across all concatenated images. In our virtual try-on pipeline, the person image resolution is 1024×768, whereas the mask and pose conditions contain substantial information redundancy and do not require such high fidelity. 

Inspired by OminiControl2~\cite{tan2025ominicontrol2efficientconditioningdiffusion}, we apply 2x downsampling along height and width to the mask and pose conditions in pixel space, thereby reducing their token count to 25\% in latent space. 

\begin{figure}[tb]
    \centering
    \includegraphics[width=\linewidth]{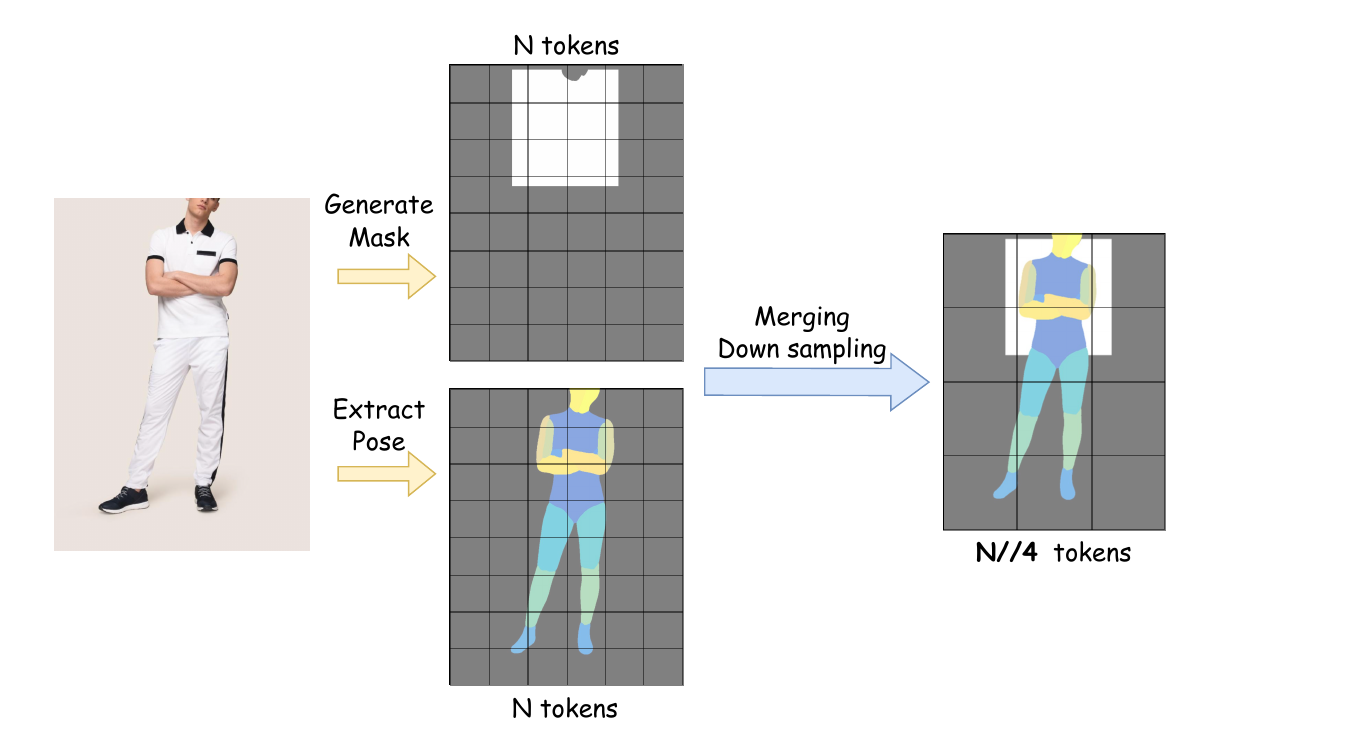}
    \vspace{-8mm}
    \caption{Our method for efficient spatial condition injection. We directly paste the pose condition onto the agnostic image, then perform downsampling, eventually to reduce the 2N tokens to N/4 tokens, achieving 87.5\% token reduction.}
    \vspace{-1mm}
    \label{fig:spatial_merging}
\end{figure}

We further merge the pose condition onto the mask condition. As shown in \cref{fig:spatial_merging}, this hierarchical compression reduces the final token count to merely 12.5\% of the original dual-condition representation, resulting in improvements in both training and inference efficiency.
\subsubsection{Region-Aware Loss Weighting}

In virtual try-on tasks, models often face a trade-off between reconstructing backgrounds and garment textures, where excessive focus on backgrounds may distract from clothing details.

We leverage human parsing results to distinguish body and background regions. The parsing mask is downsampled by 16× to match latent space resolution, constructing a region-aware weight map: body regions receive weight $1+\lambda$, backgrounds receive $1-\lambda$. This strategy enables the model to focus on garment details while reducing background interference.
\begin{equation}
  \mathbf{M}_{\text{latent}} = \text{Downsample}(\mathbf{M}_{\text{parsing}}, s=16)
\end{equation}
\begin{equation}
  \mathbf{W} = \mathbf{M}_{\text{latent}} \cdot (1+\lambda) + (1-\mathbf{M}_{\text{latent}}) \cdot (1-\lambda)
\end{equation}
\begin{equation}
  \mathcal{L}_{\text{weighted}} = \mathbb{E}_{t,\mathbf{z}_0,\boldsymbol{\epsilon}} \left[ \mathbf{W} \odot \|\boldsymbol{v} - \boldsymbol{v}_\theta(\mathbf{z}_t, t, \mathbf{c})\|^2 \right]
\end{equation}
where $\mathbf{M}_{\text{parsing}}$ is the binary parsing mask, $\mathbf{W}$ is the weight map, $\lambda \in [0,1)$ controls weighting intensity, and $\odot$ denotes element-wise multiplication.

\subsection{Promptable Dressing Style}
\label{sec:promptable}

\begin{figure}[tb]
    \centering
    \includegraphics[width=1\linewidth]{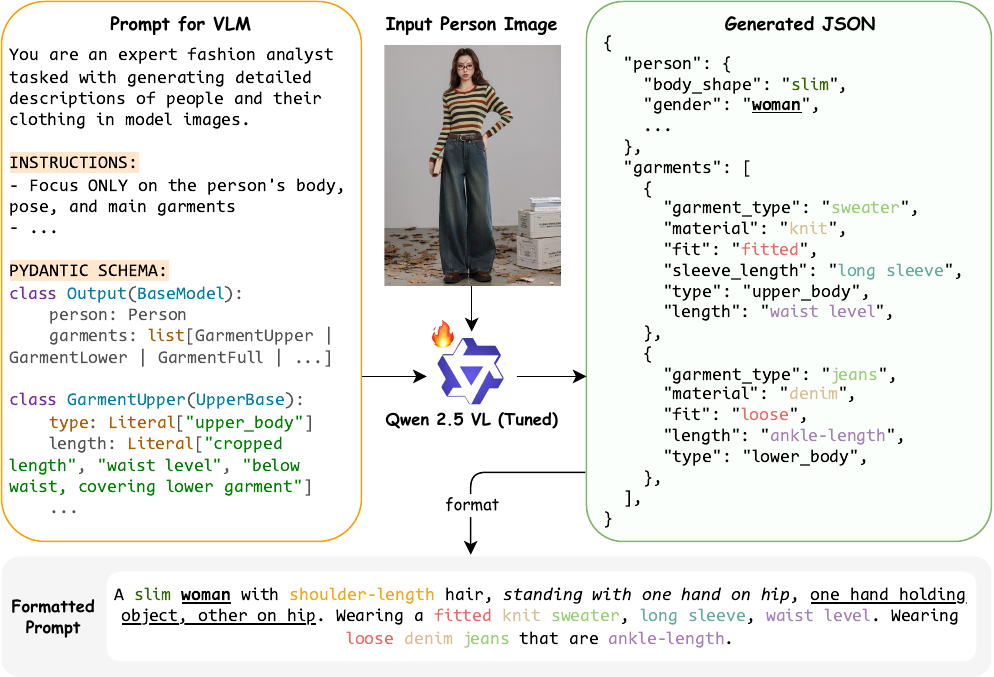}
    \vspace{-6mm}
    \caption{Overview of the \textbf{Style Prompt System}. A finetuned VLM processes an image of a person wearing the source garment, extracting person and garment attributes into a JSON format. This structured JSON is then converted into a natural language prompt for our pipeline.}
    \vspace{-5mm}
    \label{fig:prompt}
\end{figure}

Corresponding images of models wearing garments contain valuable information about wearing styles and garment-body spatial relationships. Therefore, unlike FiTDiT~\cite{jiang2024fitditadvancingauthenticgarment} that discards text control, we retain the text embedding component.

PromptDresser~\cite{kim2024promptdresserimprovingqualitycontrollability} introduced a prompt extraction pipeline using GPT-4o~\cite{openai2024gpt4technicalreport}. However, we identify several drawbacks: 1) its reliance on few-shot examples leads to substantial token consumption; 2) our inspection revealed numerous extraction inaccuracies; and 3) it is limited to extracting attributes for only a single garment. This last limitation is critical, as describing all other garments is crucial for maintaining outfit consistency.

To address these issues, we designed a new pipeline. First, we created a comprehensive, multi-garment JSON schema and used Pydantic to generate an OpenAPI-compliant specification, which we found LLMs parse more reliably than natural language descriptions. Second, to reduce cost and improve accuracy, our pipeline uses only the single person image as input. We employ a two-stage strategy: a large Qwen2.5-VL-72B~\cite{bai2025qwen25vltechnicalreport} model annotates a small dataset, which we then strictly filter to remove all non-compliant annotations. This cleaned dataset is used to finetune a much smaller Qwen2.5-VL-7B model. The resulting 7B model is not only significantly faster but also achieves higher accuracy than the 72B model, as it was trained exclusively on valid, schema-compliant data.

% We first employ the Qwen-VL-72B~\cite{bai2025qwen25vltechnicalreport} model to annotate full-body outfit attributes of human models, including sleeve length, sleeve rooling style and degree of looseness. We distill these annotations to fine-tune the Qwen-VL-7B model, yielding a lightweight caption model to be our \textbf{Style Prompt System} shown in \cref{fig:prompt}

% To mitigate the interference from irrelevant clothing descriptions in the full-body model captions on the target garment generation, our pipeline generates natural language descriptions following predefined templates, and our model takes both description and agnostic mask as reference, generating the corresponding area, keeping the rest remained. In practical applications, users can leave descriptions empty or modify style attributes, enabling precise yet flexible and controllable garment generation results.
\subsection{Temporal Self-Reference for Group Conditions}
To enable efficient multi-condition integration in FLUX~\cite{flux2024}, we introduce Temporal Self-Reference, a parameter-free mechanism using 3D-RoPE for scalable condition grouping and injection.

\subsubsection{Temporal Self-Reference}
\begin{figure}[!h]
    \centering
    \vspace{-3mm}
    \includegraphics[width=0.60\linewidth]{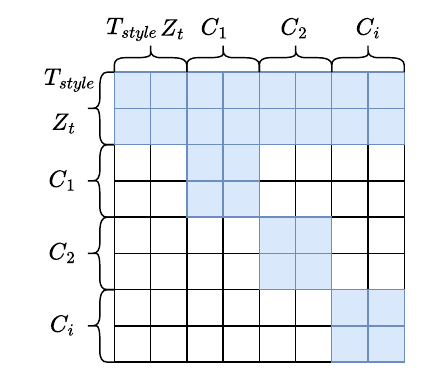}
    \vspace{-2mm}
    \caption{Attention mechanism: $z_t$ and $T_{\text{style}}$ have global visibility, while $C_i$ only attends to itself.}
    \label{fig:cache}
    \vspace{-2mm}
\end{figure}

Traditional reference networks require doubled parameters~\cite{jiang2024fitditadvancingauthenticgarment}. FastFit~\cite{chong2025fastfitacceleratingmultireferencevirtual} introduces temporal self-reference on UNet~\cite{ronneberger2015unetconvolutionalnetworksbiomedical} to avoid this. We extend it to diffusion transformers with reduced spatial condition overhead.

\begin{figure}[!h]
    \centering
    \includegraphics[width=0.85\linewidth]{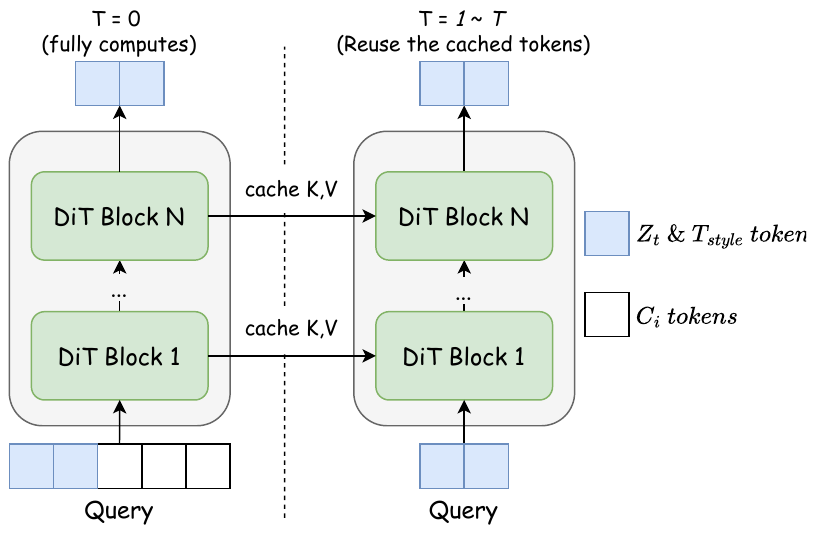}
    \vspace{-3mm}
    \caption{Temporal Self-Reference mechanism.}
    \vspace{-5mm}
    \label{fig:self-reference}
\end{figure}

As shown in \cref{fig:cache}, each $C_i$ self-attends, while $T_{\text{style}}$ and $z_t$ have global visibility. At inference (\cref{fig:self-reference}), we cache $C_i$ Key-Value pairs at the first timestep and reuse them with Query containing only $T_{\text{style}}$ and $z_t$:
\begin{equation}
Q = [Q_{z_t}, Q_{T_{\text{style}}}]
\end{equation}
\begin{equation}
K = [K_{z_t}, K_{T_{\text{style}}}, K_{C_1}^{\text{cached}}, \ldots, K_{C_n}^{\text{cached}}]
\end{equation}
\begin{equation}
V = [V_{z_t}, V_{T_{\text{style}}}, V_{C_1}^{\text{cached}}, V_{C_2}^{\text{cached}}, \ldots, V_{C_n}^{\text{cached}}]
\end{equation}

This achieves nearly lossless quality without reference network duplication.

\subsubsection{Group Conditioning by 3D-RoPE}

Building upon OminiControl~\cite{tan2025ominicontrol} and FLUX Kontext~\cite{labs2025flux1kontextflowmatching}, we repurpose RoPE's~\cite{su2023roformerenhancedtransformerrotary} temporal dimension as group condition identifiers, distinguishing multiple conditions via position encodings:
\begin{equation}
(t, x, y)_{Z_t} = (0, x_{z_t}, y_{z_t})
\end{equation}
\begin{equation}
(t, x, y)_{\mathcal{C}_i} = 
\begin{cases}
(i, x, y)_{Z_t} & \text{for spatial conditions} \\
(i, x, y + \Delta)_{Z_t} & \text{for garment conditions}
\end{cases}
\end{equation}

This parameter-free approach enables \textbf{single-garment} training and \textbf{multi-garment} inference in a \textbf{single forward pass}, avoiding iterative error accumulation.

\section{Experiment}
\begin{figure*}[!htb]
    \centering
    \includegraphics[width=0.75\linewidth]{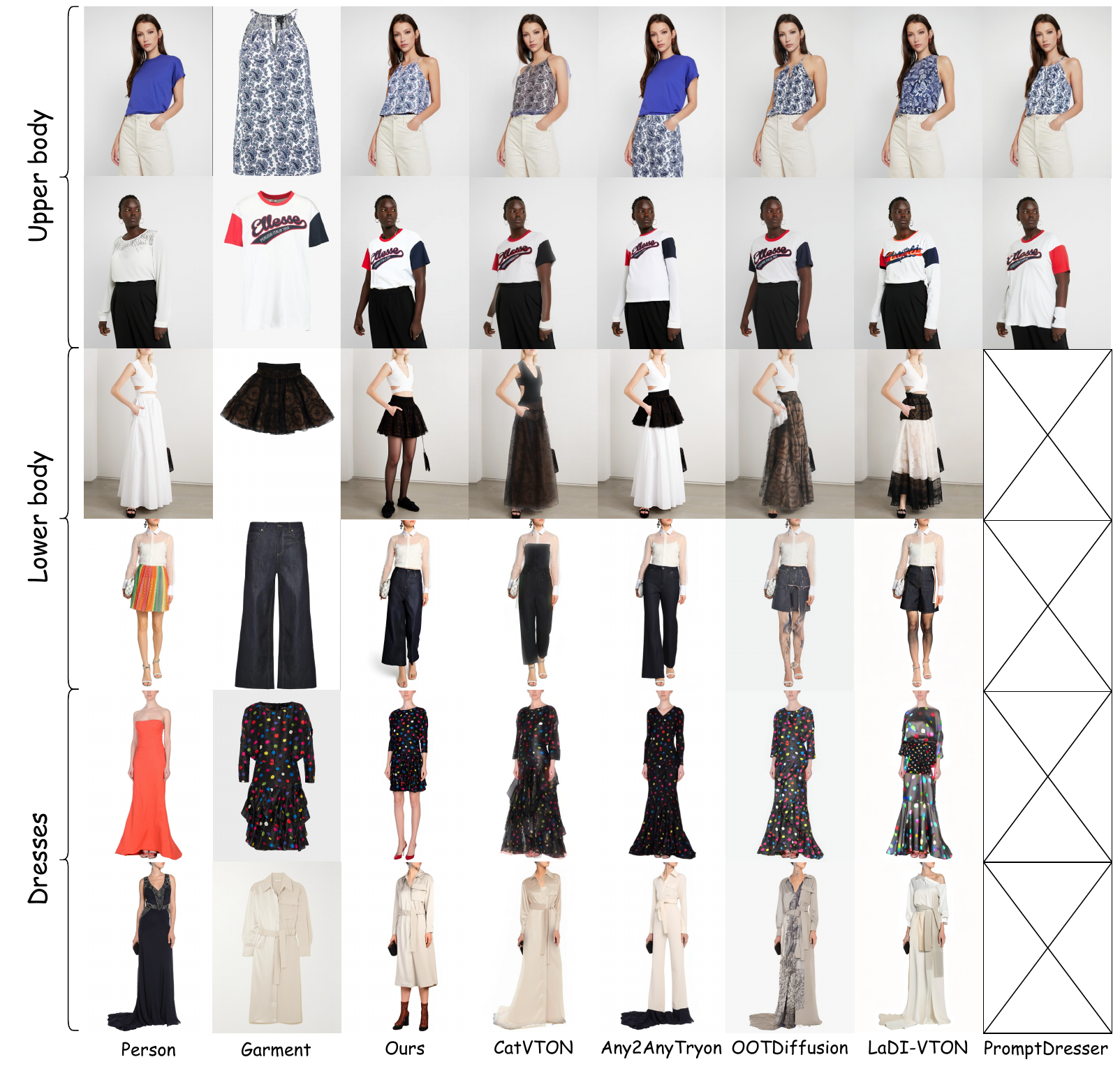} 
    \vspace{-3.5mm}
    \caption{Visualization of comparison between ours and baseline models on DressCode dataset. Best viewed when zoomed in.}
    \vspace{-4mm}
    \label{fig:main_compare}
\end{figure*}
\subsection{Dataset}
Our experiments are conducted on two public datasets and one self-collected in-the-wild dataset. VITON-HD~\cite{choi2021vitonhdhighresolutionvirtualtryon} comprises 13,679 image pairs of upper-body garments, with 2,032 pairs designated for testing. DressCode~\cite{morelli2022dresscodehighresolutionmulticategory} contains 53,792 image pairs spanning three categories (upper-body, lower-body, and dresses), including 5,400 pairs reserved for evaluation. For the in-the-wild dataset, we curate 13 human model images and 40 garment images, generating 520 image pairs through exhaustive combinations. For all test datasets, we employ our self-trained DensePose model.
\subsection{Implemetation Detail}
We build our model upon the FLUX.1-dev version~\cite{labs2025flux1kontextflowmatching}, employing Low-Rank Adaptation (LoRA)~\cite{hu2021loralowrankadaptationlarge} to efficiently fine-tune all linear layers in the Diffusion Transformer component. With a LoRA rank of 128, this introduces 580M trainable parameters. $\lambda$ for parsing loss is set to 0.5. We adopt the Prodigy~\cite{mishchenko2024prodigyexpeditiouslyadaptiveparameterfree} optimizer with its default learning rate of 1.

All experiments are conducted on the training splits of VITON-HD~\cite{choi2021vitonhdhighresolutionvirtualtryon} and DressCode~\cite{morelli2022dresscodehighresolutionmulticategory} datasets at a resolution of 1024×768. Training is performed on 16 NVIDIA H800 GPUs with an effective batch size of 16 for 90,000 steps.

\subsection{Metrics}
% 作为在我们的\modelname与先前的虚拟试衣的方法以及开源图像编辑模型的对比，使用了SSIM、LPIPS、FID，KID来衡量paired数据；使用FID,KID 来衡量unpaired数据，这些指标都是在VITON_HD, DressCode上进行的。
% 在in the wild 数据集，我们与目前市面上的一些专用的虚拟试衣模型进行了主观测评上的比较，并且以in the wild 数据集作为基准计算了FID, KID的指标
We evaluate our \modelname by comparing it with prior virtual try-on approaches and image editing models. For the quantitative evaluation on VITON-HD and DressCode datasets, we utilize SSIM~\cite{SSIM}, LPIPS~\cite{lpips}, FID~\cite{heusel2017gansfid}, and KID~\cite{bińkowski2021demystifyingmmdganskid} metrics for paired data, while employing FID and KID for unpaired data. Additionally, we conduct subjective evaluations through user studies on in-the-wild datasets, comparing our method against state-of-the-art commercial virtual try-on models.
\begin{figure*}[!htb]
    \centering
    \includegraphics[width=0.75\linewidth]{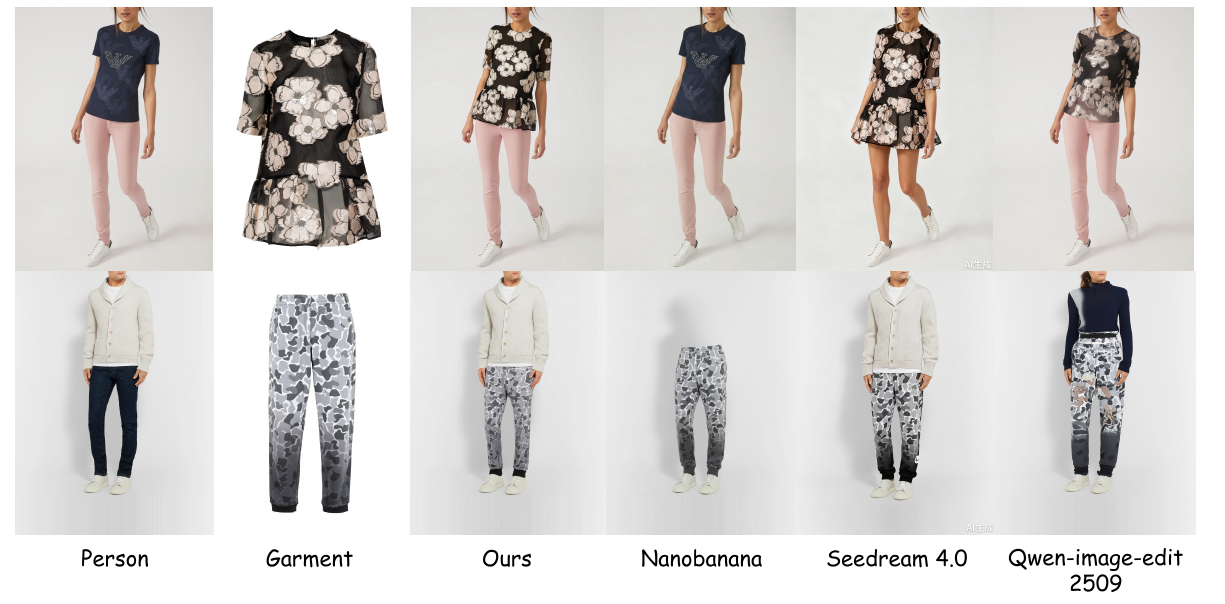} 
    \vspace{-3.5mm}
    \caption{Visualization of comparison on virtual try-on between our \modelname between state-of-the-art image-editing model on DressCode dataset. Best viewed when zoomed in.}
    \vspace{-3mm}
    \label{fig:edit_compare}
\end{figure*}
\subsection{Qualitative Comparison}
\label{subsec:qualitative}
% Figure X中，展示了虚拟试衣在开源数据集VITON-HD以及VITON数据集的可视化结果。可以看到，其他方法存在一些artifact，而我们的方法能更加完整，自然的将garment的细节进行上身,并且保留了文字和纹理的细节
\cref{fig:main_compare} presents visualization results of virtual try-on on the open-source VITON-HD and DressCode datasets. As shown, other methods exhibit various artifacts, while our approach more completely and naturally transfers garment details onto the person, preserving intricate textures and text patterns with high fidelity.

% Figure X1 中对比了Seedream 4.0, Qwen Image Edit 2509, Geini 2.5-Flash-Image(Nanobanana)在VITON-HD, DressCode上的可视化结果，可以看到通用图像编辑模型在虚拟试衣任务中存在颜色错误，artifact明显的情况，而我们的模型能够生成的更加自然
\cref{fig:edit_compare} compares our method with general-purpose image editing models, including Seedream 4.0~\cite{seedream2025seedream40nextgenerationmultimodal}, Qwen Image Edit 2509~\cite{wu2025qwenimagetechnicalreport}, and Nanobanana (Gemini 2.5-Flash-Image), on VITON-HD and DressCode datasets. The results demonstrate that general image editing models suffer from color inconsistencies and prominent artifacts in virtual try-on tasks, whereas our model generates more natural and realistic results.
% Figure X2 展示了先前的虚拟试衣方法与我们的方法在In the wild 数据集上的可视化结果，对于一些体型较为壮硕或者衣物宽松的human model，我们的模型能够更加准确的估计其体型，在生成结果中保持身材一致性，减少因体型估计错误而产生的artifact。

\subsection{Quantitative Comparison}
\subsubsection{On Open-source Dataset}
% 我们在VITON-HD, Dresscode数据集的paired, unpaired子集中，进行了量化的实验计算，总体来说，我们的模型效果超越了existing的state-of-the-art方法，可见table x
We conduct quantitative experiments on both paired and unpaired subsets of VITON-HD~\cite{choi2021vitonhdhighresolutionvirtualtryon} and DressCode~\cite{morelli2022dresscodehighresolutionmulticategory} datasets. Overall, our model surpasses existing state-of-the-art methods, as demonstrated in \cref{tab:main_results}. 

%复制代码
\begin{table*}[!htbp]
    \centering
    \resizebox{0.85\textwidth}{!}{%
    \begin{tabular}{lcccccccccccc}
        \toprule 
        \multirow{2}{*}{Methods} & \multicolumn{6}{c}{DressCode} & \multicolumn{6}{c}{VITON-HD} \\
        \cmidrule(lr){2-7} \cmidrule(lr){8-13}
        & \multicolumn{4}{c}{Paired} & \multicolumn{2}{c}{Unpaired} & \multicolumn{4}{c}{Paired} & \multicolumn{2}{c}{Unpaired} \\
        \cmidrule(lr){2-5} \cmidrule(lr){6-7} \cmidrule(lr){8-11} \cmidrule(lr){12-13}
        & SSIM $\uparrow$ & LPIPS $\downarrow$ & FID $\downarrow$ & KID $\downarrow$  & FID $\downarrow$ & KID $\downarrow$ & SSIM $\uparrow$ & LPIPS $\downarrow$  & FID $\downarrow$ & KID $\downarrow$ & FID $\downarrow$ & KID $\downarrow$ \\
        \midrule
        LaDI-VTON\cite{morelli2023ladi} & 0.7564 & 0.3800 & 5.4700 & 1.9270 & 7.3475 & \underline{2.4690} & \textbf{0.8721} & 0.1531 & \textbf{6.8500} & \textbf{1.3766} & \textbf{9.6131} & \underline{2.0207} \\
        CatVTON\cite{chong2024catvtonconcatenationneedvirtual} & \underline{0.8944} & 0.1600 & 6.5372 & 3.9591 & 8.4567 & 4.4897 & 0.8673 & 0.1882 & 9.4412 & 4.7389 & 12.5716 & 5.4555 \\
        OOTDiffusion\cite{xu2024ootdiffusion} & 0.8883 & \textbf{0.0800} & 3.6623 & \underline{0.8550} & 7.0463 & 2.7910 & 0.7919 & 0.1905 & 32.8944 & 20.0843 & 39.9626 & 26.0686 \\
        Any2AnyTryon\cite{guo2025any2anytryonleveragingadaptiveposition} & \textbf{0.9107} & 0.1208 & \textbf{3.0828} & 1.0565 & \underline{5.5404} & \underline{1.5258} & \underline{0.8705} & 0.1569 & 7.1238 & 2.1797 & 10.3120 & 2.7086 \\
        PromptDresser\cite{kim2024promptdresserimprovingqualitycontrollability} & - & - & - & - & - & - & 0.8546 & \underline{0.1163} & 7.4172 & 1.5874 & 10.4356 & 2.0531 \\
        \midrule
        {\modelname (Ours)} & 0.8913 & \underline{0.0887} & \underline{3.3103} & \textbf{0.4902} & \textbf{4.7393} & \textbf{0.4992} & 0.8619 & \textbf{0.1111} & \underline{6.8903} & \underline{1.4895} & \underline{9.9034} & \textbf{1.9174} \\
        \bottomrule
    \end{tabular}
}
    \vspace{-2mm}
    \caption{Quantitative results on VITON-HD and DressCode datasets. We compare the metrics under both paired (model's clothing is the same as the given cloth image) and unpaired settings (model's clothing differs) with other methods. 
    }
    \label{tab:main_results}
    %\vspace{-3mm}
\end{table*}
\cref{tab:comparison_edit} presents comparisons with general image editing models metioned in \cref{subsec:qualitative}. Due to API rate limitations of closed-source models, we evaluate on a fixed-size subset for this comparison. The results show that our method outperforms image editing models across all metrics.
\begin{table*}[!htbp]
    \centering
    \resizebox{0.85\textwidth}{!}{%
    \begin{tabular}{lcccccccccccc}
        \toprule 
        \multirow{2}{*}{Methods} & \multicolumn{6}{c}{DressCode} & \multicolumn{6}{c}{VITON-HD} \\
        \cmidrule(lr){2-7} \cmidrule(lr){8-13}
        & \multicolumn{4}{c}{Paired} & \multicolumn{2}{c}{Unpaired} & \multicolumn{4}{c}{Paired} & \multicolumn{2}{c}{Unpaired} \\
        \cmidrule(lr){2-5} \cmidrule(lr){6-7} \cmidrule(lr){8-11} \cmidrule(lr){12-13}
        & SSIM $\uparrow$ & LPIPS $\downarrow$ & FID $\downarrow$ & KID $\downarrow$  & FID $\downarrow$ & KID $\downarrow$ & SSIM $\uparrow$ & LPIPS $\downarrow$  & FID $\downarrow$ & KID $\downarrow$ & FID $\downarrow$ & KID $\downarrow$ \\
        \midrule
        Qwen-Image-Edit-2509~\cite{wu2025qwenimagetechnicalreport} & 0.8370& 0.1831 &13.5464 &1.9944 &13.9163 &2.7315 &0.7790 &	0.2174&	20.6403& \textbf{0.5018} &\underline{18.7954}&	\underline{1.1485}  \\
        Seedream 4.0~\cite{seedream2025seedream40nextgenerationmultimodal} & 0.8217 &0.1816 &16.8140&7.2200 &25.4943&6.1900 &0.7600 &0.2352 &26.9076 &9.3880 &26.8181&8.4790  \\
        Nanobanana\cite{google2024gemini25flash} &\underline{0.8647} &\underline{0.1186} &\underline{10.7050} &\underline{1.0860} &\underline{10.6720}&\underline{0.7160} &\underline{0.8100}&\underline{0.1520}&\underline{19.1688}&\underline{1.1299}&20.4191&\textbf{0.4411} \\ 
        
        \midrule
        {\modelname (Ours)} & \textbf{0.8919} & \textbf{0.0885} & \textbf{8.9268} & \textbf{0.4674} & \textbf{10.1217} & \textbf{0.5763} & 
        \textbf{0.8439} & \textbf{0.1365} & \textbf{15.6658} & 2.0513 & \textbf{18.0356} & 1.5828 \\
        \bottomrule
    \end{tabular}
}
\vspace{-2mm}
    \caption{Quantitative results on VITON-HD and DressCode datasets. Comparison between our model with image-editing methods.}
    \vspace{-2mm}
    \label{tab:comparison_edit}
\end{table*}

\subsubsection{User Study on In-The-Wild Dataset}
We conducted a user study comparing our method with three commercial virtual try-on services. The evaluation used 13 person images paired with 40 in-the-wild garment images covering upper-body, lower-body, and dress categories, yielding 520 try-on results. 9 independent annotators assessed each result across four dimensions: (1) garment texture consistency, (2) body shape consistency, (3) garment style consistency, and (4) garment color consistency. A result is considered excellent if more than half of the annotators rate it as such. \cref{tab:user_study} reports the excellence rate of each method across all dimensions.
\begin{table}[htbp]
    \centering
    \resizebox{\columnwidth}{!}{%
    \begin{tabular}{lccccc}
        \toprule 
        Method & Texture $\uparrow$ & Body Shape $\uparrow$ & Style $\uparrow$ & Color $\uparrow$ & Overall $\uparrow$ \\
        \midrule
        Huiwa & \underline{94.42} & 88.85 & \underline{94.80} & \textbf{99.04} & \underline{78.85} \\
        Kling & 87.12 & \underline{93.46} & 79.87 & 96.53 & 60.19 \\
        Douyin & \textbf{96.73} & 79.04 & 85.19 & 95.77 & 61.54 \\
        \midrule
        \modelname (Ours) & 93.65 & \textbf{94.62} & \textbf{96.92} & \underline{97.88} & \textbf{84.42} \\
        \bottomrule
    \end{tabular}
    }
    \vspace{-2mm}
    \caption{Excellence rates (\%) of different methods on in-the-wild dataset. A result is considered excellent if more than half annotators rate it as such. Overall indicates the percentage of images that are excellent in all four dimensions. \textbf{Bold} denotes the best result and \underline{underline} denotes the second best.}
    \label{tab:user_study}
    \vspace{-3mm}
\end{table}

\begin{table}[!ht]
    \centering
    \resizebox{\columnwidth}{!}{%
    \begin{tabular}{lcccccc}
        \toprule 
        & \multicolumn{4}{c}{Paired} & \multicolumn{2}{c}{Unpaired} \\
        \cmidrule(lr){2-5} \cmidrule(lr){6-7}
        & SSIM $\uparrow$ & LPIPS $\downarrow$ & FID $\downarrow$ & KID $\downarrow$  & FID $\downarrow$ & KID $\downarrow$ \\
        \midrule
        - w/o parsing area loss & \underline{0.8896} & \textbf{0.0872} & \textbf{3.2847} & \underline{0.5076} & \textbf{4.6390} & 0.9535 \\
        - w/o style prompt & \underline{0.8896} & 0.0926 & 3.7178 & 0.8926 & \underline{5.3516} & \underline{0.6180} \\
        - w/o 3D-RoPE & 0.8702 & 0.1304 & 6.7293 & 1.7160 & 7.8175 & 2.2801 \\ 
        \midrule
        {\modelname (Ours)} & \textbf{0.8913} & \underline{0.0887} & \underline{3.3103} & \textbf{0.4902} & \underline{5.3516} & \textbf{0.4992} \\ 
        \bottomrule
    \end{tabular}%
    }
    \vspace{-3mm}
    \caption{Ablation study on DressCode dataset. The best and second-best results are highlighted in \textbf{bold} and \underline{underlined}, respectively.}
    \vspace{-5mm}
    \label{tab:ablation_tabel_1}
\end{table}

\subsection{Ablation Studies}
\noindent{\textbf{Parsing Area Loss}} We validate Parsing Area Loss contribution to garment detail preservation through controlled ablation experiments comparing two model variants under identical configurations, differing only in this component. As shown in \cref{fig:ablate_pe} and \cref{tab:ablation_tabel_1}, while quantitative improvements on open-source benchmarks are marginal due to their clean backgrounds, qualitative advantages become evident in challenging real-world scenarios. We present representative examples from In-The-Wild datasets featuring complex backgrounds, where the model with Parsing Area Loss achieves notably higher fidelity in reproducing fine-grained garment patterns.
\begin{figure}[!htbp]
    \centering
    \includegraphics[width=0.85\linewidth]{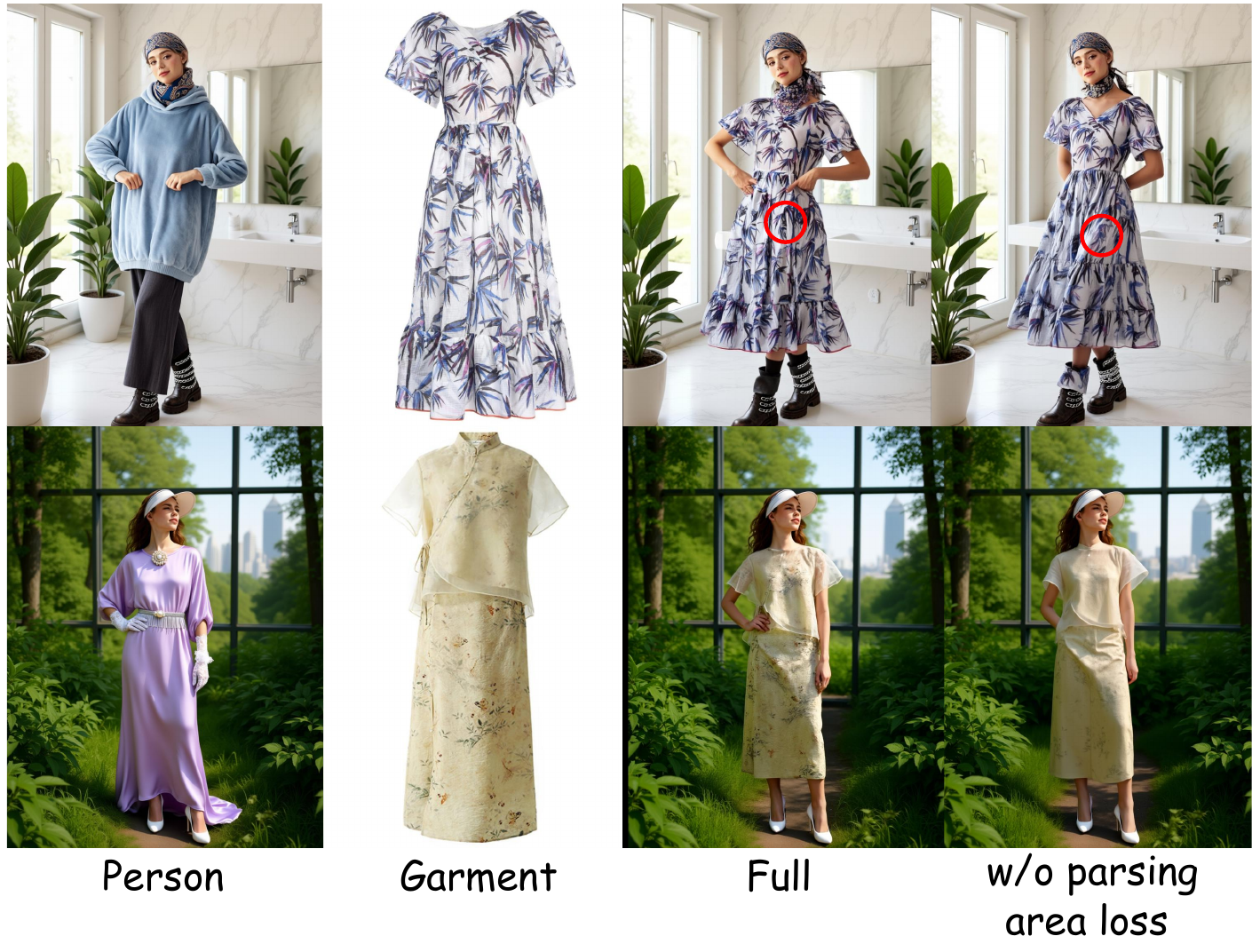}
    \vspace{-3mm}
    \caption{Visualization of ablation study on parsing area loss. Best viewed when zoomed in.}
    \vspace{-3mm}
    \label{fig:ablate_pe}
\end{figure}

\noindent{\textbf{Dressing Style Prompt}}
% 为探究文本条件对虚拟试衣效果的影响，我们设计了对照实验：在相同的测试集上分别使用null prompt和描述性prompt进行推理。尽管模型在设计上支持零文本输入的试衣生成，实验结果显示适当的文本引导仍具有积极作用。
% 如图X所示，null prompt条件下模型能够生成准确的服装着装效果；而在引入风格化描述(如"wearing [garment] in casual/formal style")后，生成结果在整体氛围、光照协调性等方面表现出进一步提升。表X的定量对比表明，prompt的引入在FID等指标上带来了边际改善，同时也为用户提供了可选的风格控制维度。
To investigate the impact of text conditions on virtual try-on performance, we design a controlled experiment: inference on the same test set using both null prompts and descriptive prompts. Although the model supports null prompt try-on generation by design, experimental results demonstrate that appropriate text guidance still yields positive effects.
Quantitative comparisons in \cref{tab:ablation_tabel_1} indicate that prompt introduction brings marginal improvements in metrics, while also providing users with an optional dimension for style control.
\begin{figure}[!htbp]
    \centering
    \vspace{-3mm}
    \includegraphics[width=0.85\linewidth]{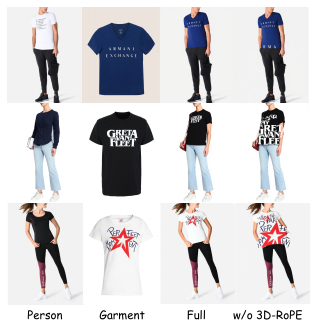}
    \vspace{-4mm}
    \caption{Visualization of ablation study on 3D-RoPE. Without 3D-RoPE, the model fails to correctly distinguish subject transfer and spatial alignment relationships between conditions.}
    \vspace{-3mm}
    \label{fig:ablate_pe}
\end{figure}

% 在我们训练\modelname的过程中，模型具备在置空Prompt情况下进行推理的能力，因此我们对比了同一个模型权重在有无Dressing Style Prompt的推理结果，可见table。
\noindent{\textbf{3D-RoPE}} To demonstrate that 3D-RoPE enhances the model's capability to distinguish  different condition groups and multiple instances within the same category, we conduct ablation experiments by removing positional encodings from all condition tokens while preserving those for text and latent tokens. As shown in \cref{fig:ablate_pe} and \cref{tab:ablation_tabel_1}, the model without 3D-RoPE fails to correctly dress the person with the target garments and produces noticeable artifacts.
% 为了证明3D-RoPE提高了模型区分不同group、同类但不同实例的条件的能力，我们将去除了所有condition 的位置编码，但保留了文本token与latent token 的位置编码，在测试集中进行了推理，结果如fig和table所示, 无3D-RoPE fail to 将衣服正确的穿搭到人物上，并产生了伪影。

\noindent{\textbf{Temporal Self-Reference.}}
% 为验证Temporal Self-Reference机制在生成质量与推理速度之间的权衡效果，我们在测试集上进行了对比实验。实验对比了启用和禁用该机制的模型在生成质量指标(FID、SSIM等)和推理速度(steps、latency)上的表现。如表X所示，Temporal Self-Reference在保持生成质量的同时显著减少了推理步数，验证了该机制在效率-质量权衡上的有效性。
To validate the effectiveness of the Temporal Self-Reference mechanism in balancing generation quality and inference speed, we conduct comparative experiments on the open-source test set. The experiments compare models with and without this mechanism across generation quality metrics and inference speed. As shown in \cref{tab:ablation_tabel_2}, Temporal Self-Reference significantly reduces inference steps while maintaining generation quality, confirming the effectiveness of this mechanism in the efficiency-quality trade-off.

\noindent{\textbf{Spatial Condition Merging.}}
% 为评估空间条件合并策略的影响，我们训练了不进行条件合并的baseline模型作为对照。两个模型在测试集上的生成速度和生成质量对比结果如表X所示。实验表明，空间条件合并在大幅降低计算开销(token数量减少至12.5\%)的同时，对生成质量的影响微乎其微，证明了该策略利用空间冗余性的合理性。
To evaluate the impact of the spatial condition merging strategy, we train a baseline model without condition merging as a control. The comparison of generation speed and quality between the two models on the test set is shown in \cref{tab:ablation_tabel_2}. Experiments demonstrate that spatial condition merging significantly reduces computational overhead while having negligible impact on generation quality, validating the rationality of exploiting spatial redundancy.
\begin{table}[tbp]
    \centering
    \resizebox{\columnwidth}{!}{%
    \begin{tabular}{lccccccc}
        \toprule 
        \multirow{2}{*}{} & \multirow{2}{*}{Inference time (s)} & \multicolumn{4}{c}{Paired} & \multicolumn{2}{c}{Unpaired} \\
        \cmidrule(lr){3-6} \cmidrule(lr){7-8}
        & & SSIM $\uparrow$ & LPIPS $\downarrow$ & FID $\downarrow$ & KID $\downarrow$  & FID $\downarrow$ & KID $\downarrow$ \\
        \midrule
        - w/o T.S.R &22.2378 & \underline{0.8913} & \underline{0.0887} & \textbf{3.3102} & 0.7975 & \textbf{4.7391} & 0.5313 \\
        - w/o S.T.M & 11.1030 & \textbf{0.8916} & \textbf{0.0873} & 3.4878 & \underline{0.5126} & 4.8546 & \textbf{0.4907} \\
        
        \midrule
        \modelname (Ours) & 9.1775 & \underline{0.8913} & \underline{0.0887} & \underline{3.3103} & \textbf{0.4902} & \underline{4.7393} & \underline{0.4992} \\
        \bottomrule
    \end{tabular}
    }
    \vspace{-1mm}
    \caption{Quantitative ablation results on DressCode, including quality metrics and inference time. \textbf{T.S.R} is for Temporal Self-Reference and \textbf{S.T.M} is for Spatial Token Merging. Inference time is averaged over 20 runs on an H800 GPU.}
    \label{tab:ablation_tabel_2}
    \vspace{-5mm}
\end{table}

\section{Conclusion}
We introduced \modelname, a novel virtual try-on framework built on a Flow Matching DiT backbone. By leveraging latent multi-modal conditional concatenation and a temporal self-reference mechanism, our model achieves a superior balance between fidelity and efficiency. \modelname supports multi-garment try-on and prompt-based style control. Experiments on VITON-HD, DressCode, and in-the-wild datasets demonstrate that our method surpasses prior VTON and general image-editing models in both visual quality and inference speed.
{
    \small
    \bibliographystyle{ieeenat_fullname}
    \bibliography{main}
}
% \input{rebuttal}

\end{document}